\documentclass{article}
\usepackage{ijcai19}

\usepackage{times}
\usepackage{soul}
\usepackage{url}
\usepackage[hidelinks]{hyperref}
\usepackage[utf8]{inputenc}
\usepackage[small]{caption}
\usepackage{graphicx}
\usepackage{amsmath}
\usepackage{booktabs}
\title{EmotionX-HSU: Adopting Pre-trained BERT for Emotion Classification}

\author{
Linkai Luo$^1$\footnote{Contact Author}\And
Yue Wang$^1$\\
\affiliations
$^1$The Hang Seng University of Hong Kong, Hong Kong, China
\emails
llk1896@gmail.com,
yuewang@hsu.edu.hk
}

\begin{document}

\maketitle

\begin{abstract}
This paper describes our approach to the EmotionX-2019, the shared task of SocialNLP 2019. To detect emotion for each utterance of two datasets from the TV show {\it Friends} and Facebook chat log {\it EmotionPush}, we propose two-step deep learning based methodology: (i) encode each of the utterance into a sequence of vectors that represent its meaning; and (ii) use a simply softmax classifier to predict one of the emotions amongst four candidates that an utterance may carry. Notice that the source of labeled utterances is not rich, we utilise a well-trained model, known as BERT, to transfer part of the knowledge learned from a large amount of corpus to our model. We then focus on fine-tuning our model until it well fits to the in-domain data. The performance of the proposed model is evaluated by micro-F1 scores, i.e., 79.1\% and 86.2\% for the testsets of {\it Friends} and {\it EmotionPush}, respectively. Our model ranks 3rd among 11 submissions.
\end{abstract}

\section{Introduction}

EmotionX-2019 shared task aims to predict one of four emotions, i.e., \textbf{neutral}, 
\textbf{joy}, \textbf{surprise}, and \textbf{anger}, for each of utterance in the dialogues of social media conversion. The datasets~\cite{DBLP:conf/lrec/HsuCKHK18} consist of scripts from a popular American TV show {\it Friends} and the anonymous Facebook chat logs named {\it EmotionPush}. We consider the task as multi-class text classification problem in the domain of natural language processing (NLP).

In recent years, deep learning technologies have been intensively used for text classification in the presence of sufficient training data. For example, convolutional neural network (CNN)~\cite{DBLP:conf/emnlp/Kim14,DBLP:conf/semeval/KimLJ18} is capable of capturing key words in the sentences, while recurrent neural network (RNN)~\cite{DBLP:conf/ijcai/LiuQH16} is proven to have better performance at modeling long dependencies among text, and so on. No matter which model is chosen, a large amount of supervised data are needed to learn the numerous model parameters. 

To tackle the low-resource problem in NLP, a lot of pre-trained models are developed for domain adoption. Pre-trained word embeddings~\cite{DBLP:conf/nips/MikolovSCCD13,DBLP:conf/emnlp/PenningtonSM14,DBLP:conf/naacl/PetersNIGCLZ18} are perhaps the most popular techniques which are frequently adopted in many NLP tasks. The generalization of word embeddings, such as sentence embeddings~\cite{DBLP:conf/emnlp/ConneauKSBB17}, are also used in downstream models, e.g., natural language inference, etc. More recently, the technique of pre-trained language models trained on large monolingual corpora combined fine-tuning has become dominant in a variety of NLP benchmarks. These models, such as BERT~\cite{DBLP:journals/corr/abs-1810-04805}, Open AI GPT~\cite{radford2018improving,radford2019language}, and Transformer-XL~\cite{DBLP:journals/corr/abs-1901-02860}, are complex neutral networks with extremely deep depth. The borrowed knowledge from pre-trained models can not only improve predictive performance for downstream models, but also accelerate model training for low-resource problems.

In this work, we adopt pre-trained BERT to our model, and focus on fine-tuning a standard feed-forward and softmax layer built on top of the pre-trained BERT. We analyze the fine-tuning methods based on the pre-trained model for the target task, and make use of the additional data generated through bidirectional translations. Our finding is that the pre-trained model can significantly improve the accuracy with only a few training epochs. We also provide a data selection strategy for imbalanced problem and achieve an significant gain compared to the prior work~\cite{DBLP:conf/acl-socialnlp/LuoYC18}.

\section{Model}

BERT is a language model built upon bidirectional training a popular attention model, Transformer~\cite{DBLP:conf/nips/VaswaniSPUJGKP17}. According to Google release, there are several model variants based on model configurations. In our work, we adopt BERT-base as the base model, which consists of an encoder with 12-layer Transformer blocks. For each block in the encoder, it contains 12-head self-attention layer and 768-dimensional hidden layer, yielding a total of 110M parameters. The base model allows inputs up to a sequence of 512 tokens and outputs the vector representations. In addition, BERT implements a technique called masked language model, that is, for each of the input sentences, it always inserts a special classification token \texttt{[CLS]} at the beginning and a special separation token \texttt{[SEP]} at the tail, and randomly chooses some words which are replaced by \texttt{[MASK]} in the sequence. More on the model description about the BERT is referred to~\cite{DBLP:journals/corr/abs-1810-04805}.

\begin{figure}[!hbt]
    \centering
    \includegraphics[scale=0.45]{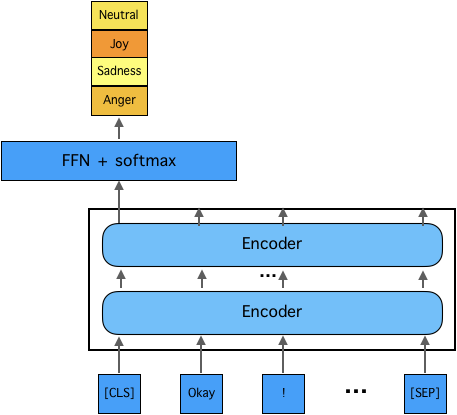}
    \caption{Pre-trained BERT for text classification. The input utterance is ``Okay!", which is labeled Joy.}
    \label{fig:bert}
\end{figure}

We use the final hidden states $\mathbf{h}$ of the first token \texttt{[CLS]} as the representation of the whole sequence $\mathbf{s}$. Interested reader can refer to~\cite{DBLP:conf/nips/VaswaniSPUJGKP17} for details about each component of the Transformer model. For our model, a standard softmax layer is then added on top of BERT to predict the probability of label $c$:
\begin{equation}
    p(c|\mathbf{s}) = \mathrm{softmax}\left(W\cdot\mathbf{h}+\mathbf{b}\right)
\end{equation}
where $W$ is the weight matrix, and $\mathbf{b}$ is the bias vector to be estimated. We then fine-tune the parameters only based on the in-domain data by maximizing the cross-entropy. The overall model architecture is shown in Figure~\ref{fig:bert}.

\section{Data}

The datasets of both \textit{Friends} and \textit{EmotionPush} are summarized in Table~\ref{tab:datasets}, and the label distributions are shown in Table~\ref{tab:labels}. A total of one thousand dialogues, or approximately 14 thousand utterances that are categorized into eight emotions, are available for both datasets. The average length (in terms of number of tokens) of the utterance is below 11; on the other hand, the maximum lengths are 91 and 227 for \textit{Friends} and \textit{EmotionPush} dataset, respectively. The labeled data are imbalanced, most of which are recognized as neutral emotion. The data labeled anger are especially insufficient compared to others for the example of \textit{EmotionPush}.

\begin{table}[!hbt]
    \centering
    \begin{tabular}{|c|c|c|c|c|c|}
        \hline
        & dialogues & utterances & tokens &avg len \\ \hline
        \textit{Friends} &1,000	&14,503	&148,043 &10.2 \\
        \hline
        \textit{EmotionPush} &1,000	&14,742	&99,073 &6.7 \\
        \hline
    \end{tabular}
    \caption{Summary of the datasets.}
    \label{tab:datasets}
\end{table}

\begin{table}[!hbt]
    \centering
    \begin{tabular}{|c|c|c|c|c|c|}
        \hline
        & neutral & joy & sadness & anger & total\\ \hline
        \textit{Friends} & 6,530 & 1,710 & 498 & 759 & 9,497\\ \hline
        \textit{EmotionPush} & 9,855 & 2,100 & 514 & 140 & 12,609 \\ \hline
    \end{tabular}
    \caption{Label distributions for the target emotions.}
    \label{tab:labels}
\end{table}

\subsection{Data Preparation}

For model training, we do not use all of the available data. We only select the data labeled with emotions of interest, i.e., neutral, joy, sadness and anger. Since additional data are provided, we also make use of them to enhance the model performance. Those data are obtained by using Google's translate API, that is, first translate the English text into another language, e.g., French, German, and Italian, and then translate it to English through a back-translate API. Although three times more data are available, we decide not to put all of them into model training. Instead, we enrich our training set by selecting the data labeled joy, sadness, and anger. The additional neutral data are discard so that the data of four classes can remain balanced.

The sentences are tokenized by default WordPiece tokenization; all the words are lowercased. We also restrict the sentence length to 128. For the sentences longer than this value, we implement a truncation method which only keeps the first 126 tokens~\footnote{two tokens are reserved for \texttt{[CLS]} and \texttt{[SEP]}.}.

\section{Fine-tuning}

For fine-tuning of the target model, we keep most of the hyperparameters the same as pre-traning in~\cite{DBLP:journals/corr/abs-1810-04805}, except for batch size, learning rate, and train epochs. We find that by choosing a small size batch, starting with a smaller learning rate, and training with a few epochs is perhaps the optimized strategy for our model. Because most of transfer learning models suffer from catastrophic forgetting problem, that is, the learnt information is erased when learning the new knowledge from target domain data. We also find that using a aggressive learning rate such as 5e-4 make the training fail to converge. To summarize, we discover that the following values work well for us:
\begin{itemize}
    \item batch size: 24
    \item learning rate (Adam): 2e-5
    \item train epochs: 4
\end{itemize}

\section{Results}

The evaluation is conducted separately on \textit{Friends} and \textit{EmotionPush} datasets. Each testset contains 240 dialogues. The test results are listed in Table~\ref{tab:result1} and Table~\ref{tab:result2}. 

\begin{table}[!hbt]
    \centering
    \begin{tabular}{ccccc}
    &             precision   & recall  & f1-score   & support \\\hline

     neutral      &0.801     &0.914     &0.854      &1035 \\
         joy      &0.854     &0.648     &0.736       &505 \\
     sadness      &0.608     &0.512     &0.556       &121 \\
       anger      &0.662     &0.638     &0.650       &141 \\\hline
   micro avg      &0.791     &0.791     &0.791      &1802 \\
   macro avg      &0.731     &0.678     &0.699      &1802 \\
weighted avg      &0.792     &0.791     &0.785      &1802 \\\hline
    \end{tabular}
    \caption{Evaluation results for \textit{Friends}}
    \label{tab:result1}
\end{table}

\begin{table}[!hbt]
    \centering
    \begin{tabular}{ccccc}
      &        precision    &recall  &f1-score   &support\\\hline

     neutral      &0.908     &0.917     &0.913      &2146\\
         joy      &0.747     &0.730     &0.738       &601\\
     sadness      &0.627     &0.627     &0.627       &110\\
       anger      &0.474     &0.333     &0.391        &27\\\hline
       
   micro avg      &0.862     &0.862     &0.862      &2884\\
   macro avg      &0.689     &0.652     &0.667      &2884\\
weighted avg      &0.860     &0.862     &0.861      &2884\\\hline
    \end{tabular}
    \caption{Evaluation results for \textit{EmotionPush}}
    \label{tab:result2}
\end{table}

In general, our model has better performance in classifying emotions for \textit{EmotionPush} than \textit{Friends}, in terms of average F1 score. We find that most of the other submissions yield the similar results, see the leaderborads~\footnote{\url{https://sites.google.com/view/emotionx2019/leaderboards?authuser=0}} for reference. We guess a possible explanation is the language they use, e.g., the language in \textit{EmotionPush} is less complex than that in \textit{Friends}. Compared to the best score submission, our major gap is the accuracy in predicting the emotion of anger, especially for \textit{EmotionPush}. Because the labeled data for some emotions is far from well learning a deep neutral network model, we may rely on extra resource in order to treat the extremely low-resource properly.

Also pay attention to the merits for each category of the emotions, it is clear that the predictive performance for each label directly relates to the amount of labeled training data, for example, when we refer to Table~\ref{tab:labels} and the recall in Table~\ref{tab:result1} and Table~\ref{tab:result2}. It implies that the imbalanced data problem still exits; however, quite interestingly, the precision of predicting emotion of joy is much higher than predicting neutral emotion for \textit{Friends} dataset.

\section{Conclusion}

We propose to use pre-trained BERT model to improve the model performance for emotion classification in the social media conversions. It is evident that the pre-training can significantly increase predictive accuracy and simultaneously boost model training with only a few training epochs. Nevertheless, the emotion classification is more than text classification, as pointed out in~\cite{DBLP:conf/acl-socialnlp/LuoYC18}, that a precise recognition of emotions for certain utterance also relates to the context it belongs to. It is straightforward that even though the same text could express different emotions according to the conversation it engages. We hope to incorporate contextual information into pre-trained model to further enhance our system.

\section*{Acknowledgments}

This work is supported by Research Grants Council of the Hong Kong Special Administrative Region, China (Project Reference No.: UGC/FDS14/E06/18).

\bibliographystyle{named}
\bibliography{ijcai19}

\end{document}